\documentclass[10pt, conference, compsocconf]{IEEEtran}
\IEEEoverridecommandlockouts
\usepackage{cite}
\usepackage{amsthm,amssymb} %,amsfonts
\usepackage{algorithmic}
\usepackage{graphicx}
\usepackage{textcomp}
\usepackage{xcolor}
\usepackage{enumitem}
\usepackage{booktabs} % For formal tables
\usepackage{multirow}
\usepackage{caption}
\usepackage{amsmath}
\usepackage{url}   
\newtheorem{theorem}{Theorem}

\def\BibTeX{{\rm B\kern-.05em{\sc i\kern-.025em b}\kern-.08em
    T\kern-.1667em\lower.7ex\hbox{E}\kern-.125emX}}
\begin{document}

\title{Deep Multi-attributed Graph Translation with Node-Edge Co-evolution}

\author{\IEEEauthorblockN{Xiaojie Guo, Liang Zhao, Cameron Nowzari, Setareh Rafatirad\\ Houman Homayoun, and Sai Manoj Pudukotai Dinakarrao}
\IEEEauthorblockA{George Mason University\\
Fairfax, VA, USA.\\
\{xguo7,lzhao9,cnowzari,srafatir,hhomayou,spudukot\}@gmu.edu}
}

\maketitle

\begin{abstract}
%As deep graph learning and generation models are utilized in more and more real world applications involving molecular generation and visual Relationship Detection.
Generalized from image and language translation, graph translation aims to generate a graph in the target domain by conditioning an input graph in the source domain. This promising topic has attracted fast-increasing attentions recently. Existing works are limited to either merely predicting the node attributes of graphs with fixed topology or predicting only the graph topology without considering
node attributes, but cannot simultaneously predict both of them, due to substantial challenges: 1) difficulty in characterizing the interactive, iterative, and asynchronous translation process of both nodes and edges and 2)~difficulty in discovering and maintaining the inherent consistency between the node and edge in predicted graphs. These challenges prevent a generic, end-to-end framework for joint node and edge attributes prediction, which is a need for real-world applications such as malware confinement in IoT networks and structural-to-functional network translation. These real-world applications highly depend on hand-crafting and ad-hoc heuristic models, but cannot sufficiently utilize massive historical data. In this paper, we termed this generic problem ``multi-attributed graph translation" and developed a novel framework integrating both node and edge translations seamlessly. The novel edge translation path is generic, which is proven to be a generalization of the existing topology translation models. Then, a spectral graph regularization based on our non-parametric graph Laplacian is proposed in order to learn and maintain the consistency of the predicted nodes and edges. Finally, extensive experiments on both synthetic and real-world application data demonstrated the effectiveness of the proposed method.
\end{abstract}


\section{Introduction}
Many problems regarding structured predictions are encountered in the process of "translating" an input data (e.g., images, texts) into a corresponding output data, which is to learn a translation mapping from the input domain to the target domain. For example, many problems in image processing and computer vision can be seen as a "translation" from an input image into a corresponding output image. Similar applications can also be found in language translation~\cite{xu2018graph2seq,xu2018exploiting,xu2018sql}, where sentences (sequences of words) in one language are translated into corresponding sentences in another language. Such generic translation problem, which is important yet has been extremely difficult in nature, has attracted rapidly-increasing attention in recent years. The conventional data translation problem typically considers the data under special topology. For example, an image is a type of grid where each pixel is a node and each node has connections to its spatial neighbors. Texts are typically considered as sequences where each node is a word and an edge exists between two contextual words. Both grids and sequences are special types of graphs. In many practical applications, it is required to work on data with more flexible structures than grids and sequences, and hence more powerful translation techniques are required in order to handle more generic graph-structured data. This has been widely applied into many applications, e.g. predicting future states of a system in the physical domain based on the fixed relations (e.g. gravitational forces) among nodes~\cite{battaglia2016interaction} and the traffic speed forecasting on the road networks~\cite{li2017diffusion,yu2017spatio}. Though they can work on generic graph-structured data, they assume that the graphs from the input domain and target domain share the same graph topology but cannot model or predict the change of the graph topology.

To address the above issues where the topology can change during translation, deep learning-based graph translation problem has debuted in the very recent years. This problem is promising and critical to the domains where the variations of the graph topology are possible and frequent such as social network and cyber-network. For example, in social networks where people are the nodes and their contacts are the edges, the contact graph among them vary dramatically across different situations. For example, when the people are organizing a riot, it is expected that the contact graph to become denser and several special ``hubs'' (e.g., key players) may appear. Hence, accurately predicting the contact network in a target situation is highly beneficial to situational awareness and resource allocation. Existing topology translation models~\cite{guo2018deep, sun2019graph} predict the graph topology (i.e., edges) in a target domain based on that in an input domain. They focus on predicting the graph topology but assume that the node attributes value are fixed or do not exist.
\begin{figure}[htb]
\centering
%\captionsetup{font={small}}
\includegraphics[width=0.35\textwidth]{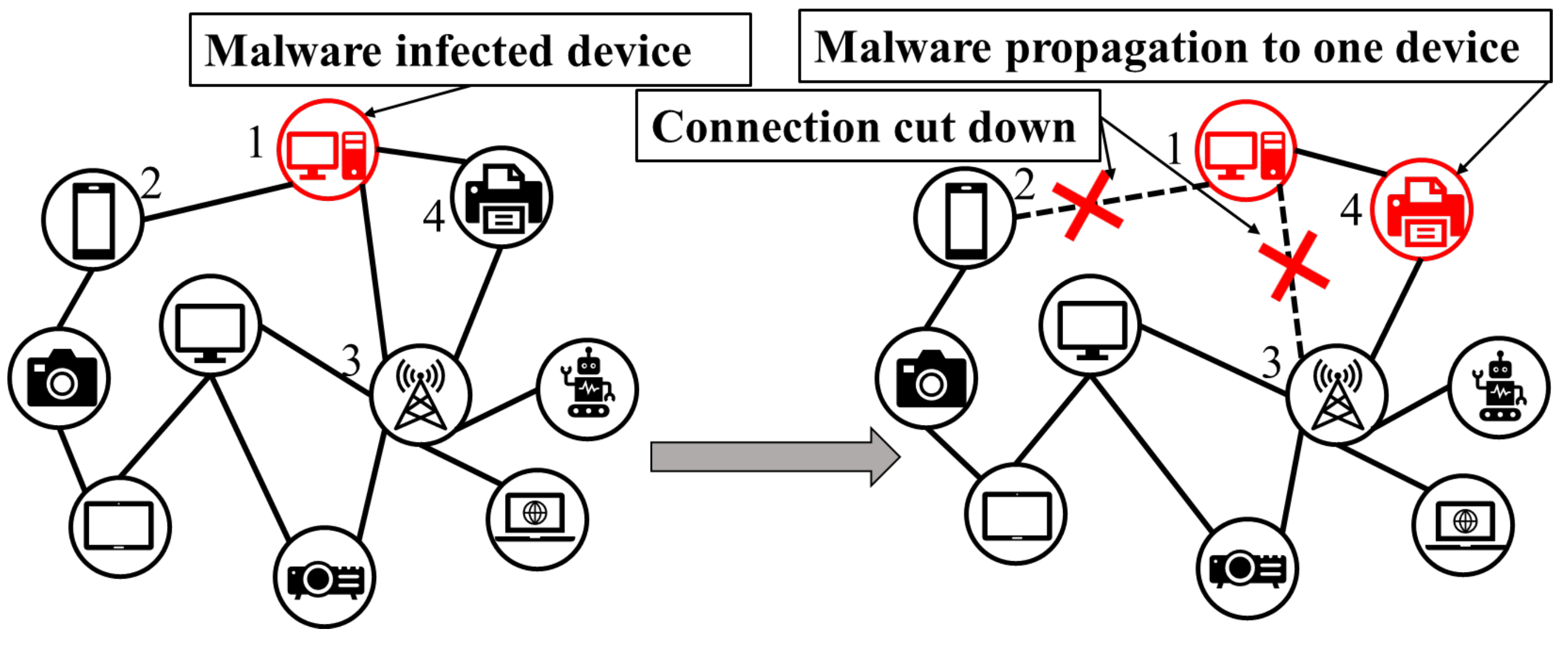}
\caption{Given the network at time $t$ (shown in the left graph), malware confinement is conducted to predict the most optimal status at time $t+\gamma$ shown in the right, where Devices 2 and 3 are protected by cutting the links (edges) to the compromised Device 1, while the Device 4 is propagated by malware without cutting link.}
\label{fig:example}
\end{figure}

Therefore, existing works either predict node attributes upon fixed topology or predict edge attributes upon fixed node attributes. However, in many applications, both node attributes and edge attributes can change. In this paper, such generic problem is named as \emph{multi-attributed graph translation}, with important real-world applications ranging from biological structural to functional network translation~\cite{abdelnour2018functional} to network intervention research~\cite{sayadi2018ensemble}. For example, the process of malware confinement\footnote{A device infected in an IoT network can propagate to other nodes connected to it, leading to contaminating the whole network, such as MiraiBot attack. As such, it is non-trivial to confine the malware to limit the infection and also equally important to maintain overall network connectivity and performance.} over IoT (Internet of Things) is typically a graph translation problem as shown in  Fig.~\ref{fig:example}. It takes the initial status of IoT as input, and predicts the target graph which is ideally the optimal status of the network with modified connections (i.e., edges) and devices (i.e., nodes) state that helps to limit malware propagation and maintain network throughput. Epidemic controlling can also be considered as a multi-attributed graph translation problem, which is to estimate how the initial disease contact network (i.e., multi-attributed edges) and the human health stage (i.e., multi-attribute nodes) are jointly changed after the specific interventions. Since multi-attributed graph translation problem is highly sophisticated, there is no generic framework yet, but only ad-hoc methods for few specific domains, which heavily rely on intensive hand-crafting and domain-specific mechanistic models that could be extremely time- and resource- consuming to run in large scale. Hence, a generic, efficient, and end-to-end framework for general multi-attributed graph translation problems is highly in demand. Such framework needs to be able to comprehensively learn the translation mapping, remedy human bias by enjoying the large historical data, and achieve efficient prediction.

In this paper, we focus on the generic problem of multi-attributed graph translation, which cannot be handled by the existing methods because of the following challenges: 1)~\textbf{Translation of node and edge attributes are mutually dependent}. The translation of edge attributes should not only consider edges, but also the node attributes. For example, in Fig. \ref{fig:example}, two links are cut down since their linked Device 1 is compromised, which exemplifies the interplay between nodes and edges. Similarly, node translation also needs to jointly consider both nodes and edges, e.g., Device 4 is infected due to its link to Device 1. All the above issues need to be jointly considered but no existing works can handle. 2)~\textbf{Asynchronous and iterative changes of node and edge attributes during graph translation}. The multi-attributed graph translation process may involve a series of iterative changes in both edge and node attributes. For example in Fig.\ref{fig:example}, the translation could take several steps since the malware propagation is an iterative process from one device to the others. The links to a device may be cut (i.e., edge changes) right after it is compromised (i.e, node attribute change). These orders and dependencies of how node and edge attributes change during the translation are very important, yet difficult to be learned. 3)~\textbf{Difficulty in discovering and enforcing the correct consistency between node attributes and graph spectra}. Although the predicted node and edge attributes are two different outputs, they should be highly dependent on each other instead of being irrelevant. For example, as shown in Fig. \ref{fig:example}, the reason why Devices 2 and 3 on the right graph are not compromised is that they have no links with the compromised Device 1 anymore. It is highly challenging to learn and maintain the consistency of node and edge attributes, which are very sophisticated and domain-specific patterns.
    
%propose the model...and contributions
To the best of our knowledge, this is the first work that addresses all the above challenges and provides a generic framework for the multi-attributed graph translation problem. This paper propose an Node-Edge Co-evolving Deep Graph Translator (NEC-DGT) with novel architecture and components for joint node and edge translation. Multi-block network with novel interactive node and edge translation paths are developed to translate both node and edge attributes, while skip-connection is utilized among different blocks to allow the non-synchronicity of changes in node and edge attributes. A novel spectral graph regularization is designed to ensure the consistency of nodes and edges in generated graphs. The contributions of this work are summarized as follows:
\begin{itemize}%[leftmargin=*]
    \item\textbf{The development of a new framework for multi-attributed graph translation}. We formulate, for the first time, a multi-attributed graph translation problem and propose the NEC-DGT to tackle this problem. The proposed framework is generic for different applications where both node and edge attributes can change after translation.
    \item\textbf{The proposal of novel and generic edge translation layers and blocks}. A new edge translation path is proposed to translate the edge attributes from the input domain to the output domain. Existing edge translation methods were proven to be special cases of ours, which can handle broad multi-attribute edges and nodes.
    \item\textbf{The proposal of a spectral-based regularization that ensures consistency of the predicted nodes and edges}. In order to discover and maintain the inherent relationships between predicted nodes and edges, a new non-parametric graph Laplacian regularization with a graph frequency regularization is proposed and leveraged. 
    \item \textbf{The conduct of extensive experiments to validate the effectiveness and efficiency of the proposed model}. Extensive experiments on four synthetic and four real-world datasets demonstrated that NEC-DGT is capable of generating graphs close to ground-truth target graphs and significantly outperforms other generative models.
\end{itemize}

\section{Related Works}
\textbf{Graph neural networks learning}. In recent years, there has been a surge in research focusing on graph neural networks, which are generally divided into two categories: Graph Recurrent Networks~ \cite{gori2005new,scarselli2008graph,li2015gated} and Graph Convolutional Networks~\cite{niepert2016learning,mousavi2017hierarchical,defferrard2016convolutional,kawahara2017brainnetcnn,nikolentzos2017kernel,cao2016deep,kipf2016semi,wu2019scalable}. Graph Recurrent Networks originates from the early works of graph neural networks proposed by Gori et al.~\cite{gori2005new} and Scarselli et al.~\cite{scarselli2008graph} based on recursive neural networks. Another line of research is to generalize convolutional neural networks from grids (e.g., images) to generic graphs. Bruna et al.~\cite{bruna2013spectral} first introduced the spectral graph convolutional neural networks, and then it was extended by Defferrard et al.~\cite{defferrard2016convolutional} using fast localized convolutions, which is further approximated for an efficient architecture for a semi-supervised setting~\cite{kipf2016semi}.

\textbf{Graph generation}. 
Most of the existing GNN based graph generation for general graphs have been proposed in the last two years and are based on VAE \cite{simonovsky2018graphvae,samanta2018designing} and generative adversarial nets (GANs) \cite{bojchevski2018netgan}, among others \cite{li2018learning,you2018graphrnn}. 
Most of these approaches generate nodes and edges sequentially to form a whole graph, leading to the issues of being sensitive to the generation order and very time-consuming for large graphs. 
%Similarly, Bojchevski et al. \citep{bojchevski2018netgan} also prefer to generate nodes and edges sequentially, by random walk on the graphs which also faces similar problems. 
%\cite{simonovsky2018graphvae} and \cite{samanta2018designing} both proposed new variational autoencoders for whole graph generation, though once again they are typically only able to handle very small graphs (i.e., with $\le 50$ nodes).
%and cannot scale well in both memory and runtime for large graphs. 
Differently, GraphRNN \cite{you2018graphrnn} builds an autoregressive generative model on these sequences with LSTM model and has demonstrated its good scalability.

\textbf{Graph structured data translation}. The existing Graph structured data translation either deal with the node attributes prediction or translate the graph topology. Node attributes prediction aims at predicting the node attributes given the fixed graph topology~\cite{battaglia2016interaction,li2017diffusion,yu2017spatio,gao2018local}. 
%Battaglia et al.~\cite{battaglia2016interaction} introduced the interaction network to formalize the objects interaction into graphs and reason about how objects in complex systems interact. 
Li et al.~\cite{li2017diffusion} propose a Diffusion Convolution Recurrent Neural Network (DCRNN) for traffic forecasting which incorporates both spatial and temporal dependency in the traffic flow. Yu et al.~\cite{yu2017spatio} formulated the node attributes prediction problem of graphs based on the complete convolution structures. Graph topology translation considers the change of graph topology from one domain distributions to another. 
%Kawahara et al.~\cite{kawahara2017brainnetcnn} leveraged the topological locality of structural brain networks and proposed BrainNetCNN, a convolutional neural network (CNN) framework to predict clinical neurodevelopmental outcomes from brain networks based on several graph convolution filters. 
Guo et al.~\cite{guo2018deep} proposed and tackled graph topology translation problem by proposing a generative model consisting of a graph translator with graph convolution and deconvolution layers and a new conditional graph discriminator. Sun et al.~\cite{sun2019graph} proposed a graphRNN based model which generates a graph's topology based on another graph.

\section{Problem Formulation}
\label{section: problem}
This paper focuses on predicting a \emph{target multi-attributed graph} based on an \emph{input multi-attributed graph} by learning the graph translation mapping between them. The following provides the notations and mathematical problem formulation.

\begin{table}[h]
  \centering
  \caption{Important notations and descriptions}
  \begin{tabular}{p{2cm}p{5.8cm}}\\
  \hline
    Notations& Descriptions\\
    \hline
     $G(\mathcal{V}_0,\mathcal{E}_0,E_0,F_0)$ &Input graph with node set $\mathcal{V}_0$, edge set $\mathcal{E}_0$, edge attributes tensor $E_0$ and node attributes matrix $F_0$\\
     $G(\mathcal{V}',\mathcal{E}',E',F')$ &Target graph with node set $\mathcal{V}'$, edge set $\mathcal{E}'$, edge attributes tensor $E'$ and node attributes matrix $F'$\\
     $C$ &Contextual information vector\\
     $N$ &Number of nodes\\
     $M$ &Number of edges\\
     $D$ &Dimension of node attributes\\
     $K$ &Dimension of edge attributes\\
     $c$ &Dimension of contextual information vector\\
     $S$ &Number of translation blocks\\
    \hline
  \end{tabular}
  \label{table:notations}
\end{table}

Define an input graph as $G(\mathcal{V}_0,\mathcal{E}_0,E_0,F_0)$ where $\mathcal{V}_0$ is the set of $N$ nodes, and $\mathcal{E}_0\subseteq \mathcal{V}_0 \times \mathcal{V}_0$ is the set of $M$ edges. $e_{i,j}\in\mathcal E_0$ is an edge connecting nodes $i\in\mathcal V_0$ and $j\in\mathcal V_0$. $\mathcal{E}_0$ contains all pairs of nodes while the existence of $e_{i,j}$ is reflected by its attributes. $E_0\in \mathbb R^{ N\times N\times K}$ is the edge attributes tensor, where $E_{0,i,j} \in \mathbb R^{1\times K }$ denotes the edge attributes of edge $e_{i,j}$ and $K$ is the dimension of edge attributes. $F_0\in \mathbb R^{ N\times D}$ refers to the node attribute matrix, where $F_{0,i}\in \mathbb R^{1\times D}$ is the node attributes of node $i$ and $D$ is the dimension of the node attributes. Similarly, we define the target graph as $G(\mathcal{V'},\mathcal{E'},E',F')$. Note that the target and input graphs are different both in their node attributes as well as edge attributes. Moreover, vector $C$ provides some contextual information on the translation process. Therefore, multi-attributed graph translation is defined as learning a mapping: $\mathcal{T}:G(\mathcal{V}_0, \mathcal{E}_0, E_0, F_0);C\rightarrow G(\mathcal{V',E'},E',F')$.

\begin{figure}[htb]
%\captionsetup{font={small}}
\centering
\includegraphics[width=0.45\textwidth]{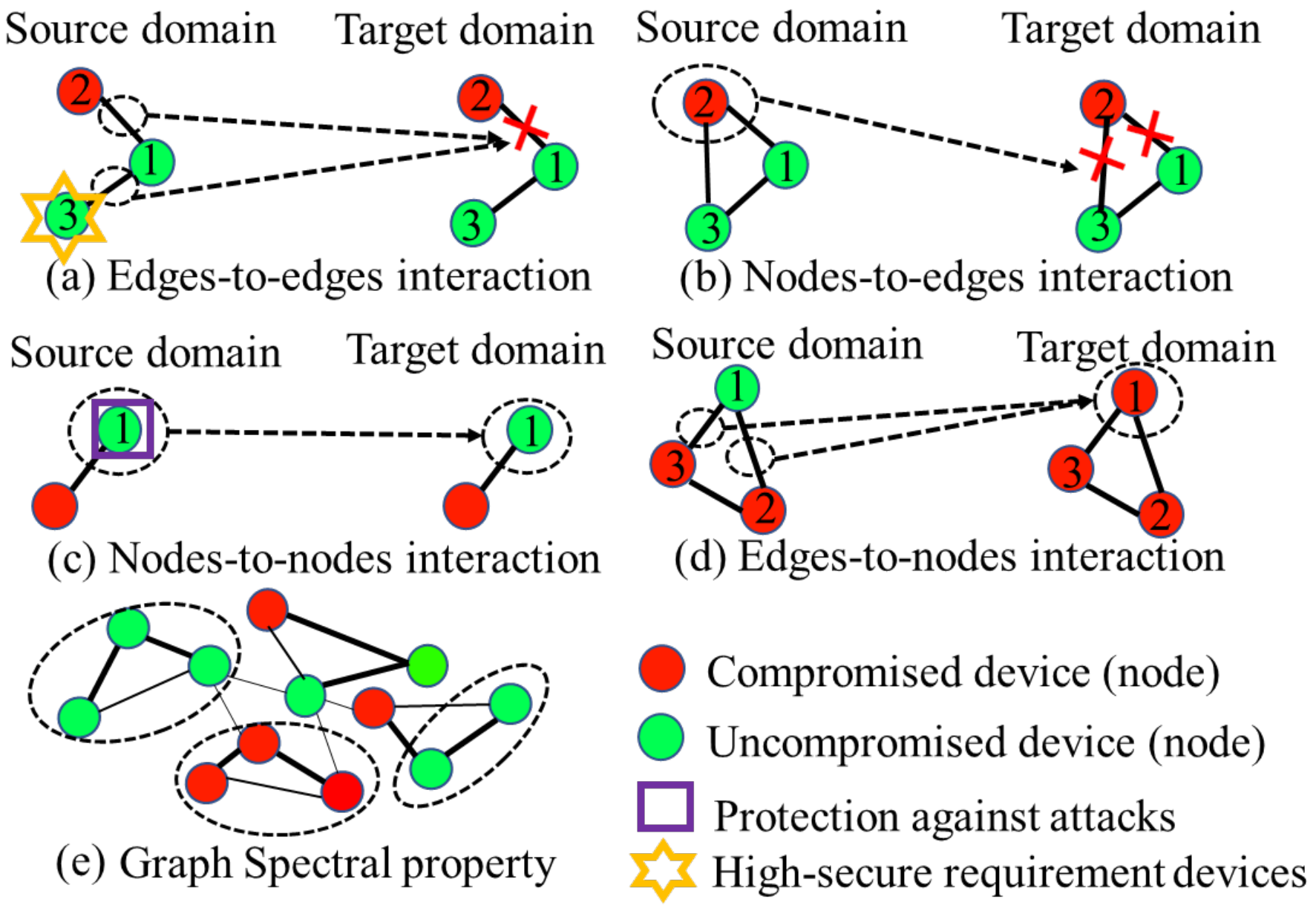}
\caption{Five types of interactions during graph translation in the example of malware confinement. Node attributes are indications of malware attacks of IoT devices and edges represent the connections between devices.}
\label{fig:interaction}
\end{figure}

For example, considering the malware confinement case where the nodes refer to IoT devices and the edges reflect the communication links between two devices. The node attributes include the malware-infection status and the properties of that device (i.e., specification and anti-virus software features). A single IoT device (i.e., node) that is compromised has the potential to spread malware infection across the network, eventually compromising the network or even ceasing the network functionality. In contrast, in order to avoid malware spreading as well as maintain the performance of the network, the network connectivity (i.e., graph topology) should be modified through \emph{malware confinement}, thus to change the device status (i.e., node attributes) accordingly. Hence, malware confinement can be considered as predicting the optimal topology as well as the corresponding node and edge attributes of the target graph, where both malware prevention and device performance are maximized.

Multi-attributed graph translation problem requires to highlight several unique considerations as depicted in Fig.\ref{fig:interaction}: 
1) \textbf{Edges-to-edges interaction}: In target domain, the edge attributes $E'_{i,j}$ of an edge $e_{i,j}$ can be influenced by its incident edges' attributes $E_{0,i,k}$ and $E_{0,k,j}$ in input domain. For example, in Fig. \ref{fig:interaction} (a), if Devices 1 and 3 must be prevented from infection, then the edges between the compromised Device 1 and Device 2 need to be cut, due to the paths among them in input domain. 
2) \textbf{Nodes-to-edges interaction}: In target domain, the attributes $E'_{i,j}$ of edge $e_{i,j}$ can be influenced by its incident nodes' attributes $F_{0,i}$ and $F_{0,j}$ in the input domain. %It means the node attributes can affect the edges' type or weights it connected to other nodes in target domain. 
As shown in  Fig. \ref{fig:interaction} (b), if Device 2 is compromised in input domain, then in target domain, only its connections to Devices 1 and 3 need to be removed but the connection between Devices 1 and 3 can be retained because they are not compromised.
3) \textbf{Nodes-to-nodes interaction}: For a given node $i$, its attribute $F_{0,i}$ in input domain may directly influence its attribute $F'_{i}$ in target domain. As shown in Fig. \ref{fig:interaction} (c), Device 3 with effective anti-virus protection (e.g. firewall) may not be easily compromised in target domain.
4) \textbf{Edges-to-nodes interaction}: For a given node $i$, its related edge attributes $E_{0,i,j}$ in input domain may affect its attributes $F'_i$ in target domain. As shown in Fig. \ref{fig:interaction} (d), Device 1 which has more connections with compromised devices in input domain is more likely to be infected in target domain. 
5) \textbf{Spectral Graph Property}: There exist relationships between nodes and edges in one graph as reflected by the graph spectrum. These relationships are claimed to have some persistent or consistent patterns across input and target domains, which have also been verified in many real-world applications such as brain networks~\cite{abdelnour2018functional}. For example, as shown in Fig. \ref{fig:interaction} (e), the devices that are densely connected as a sub-community tend to be in the same node status, which is a shared pattern for relationships between nodes and edges in different domains.

Multi-attributed graph translation should consider all the above properties, which cannot be comprehensively handled by existing methods because: 1)~Lack of a generic framework to simultaneously characterize and automatically infer all of the above node-edge interactions during translation process. 2)~Difficulty in automatically discovering and characterizing the inherent spectral relationship between the nodes and edges in each graph, and ensuring consistent spectral patterns in graphs across input and target domains. 3)~All the above interactions could be imposed repeatedly, alternately, and asynchronously during the translation process. It is difficult to discover and characterize such important yet sophisticated process.

\section{The Proposed Method: NEC-DGT}
In this section, we propose the Node-Edge Co-evolving Deep Graph Translator (NEC-DGT) to model the multi-attributed graph translation process. First, an introduction of the overall architecture and the loss functions is given. Then, the elaborations of three modules on edge translation, node translation, and graph spectral regularization are presented.
\subsection{Overall architecture}
\begin{figure}[t]
%\captionsetup{font={small}}
\centering
\includegraphics[width=\columnwidth]{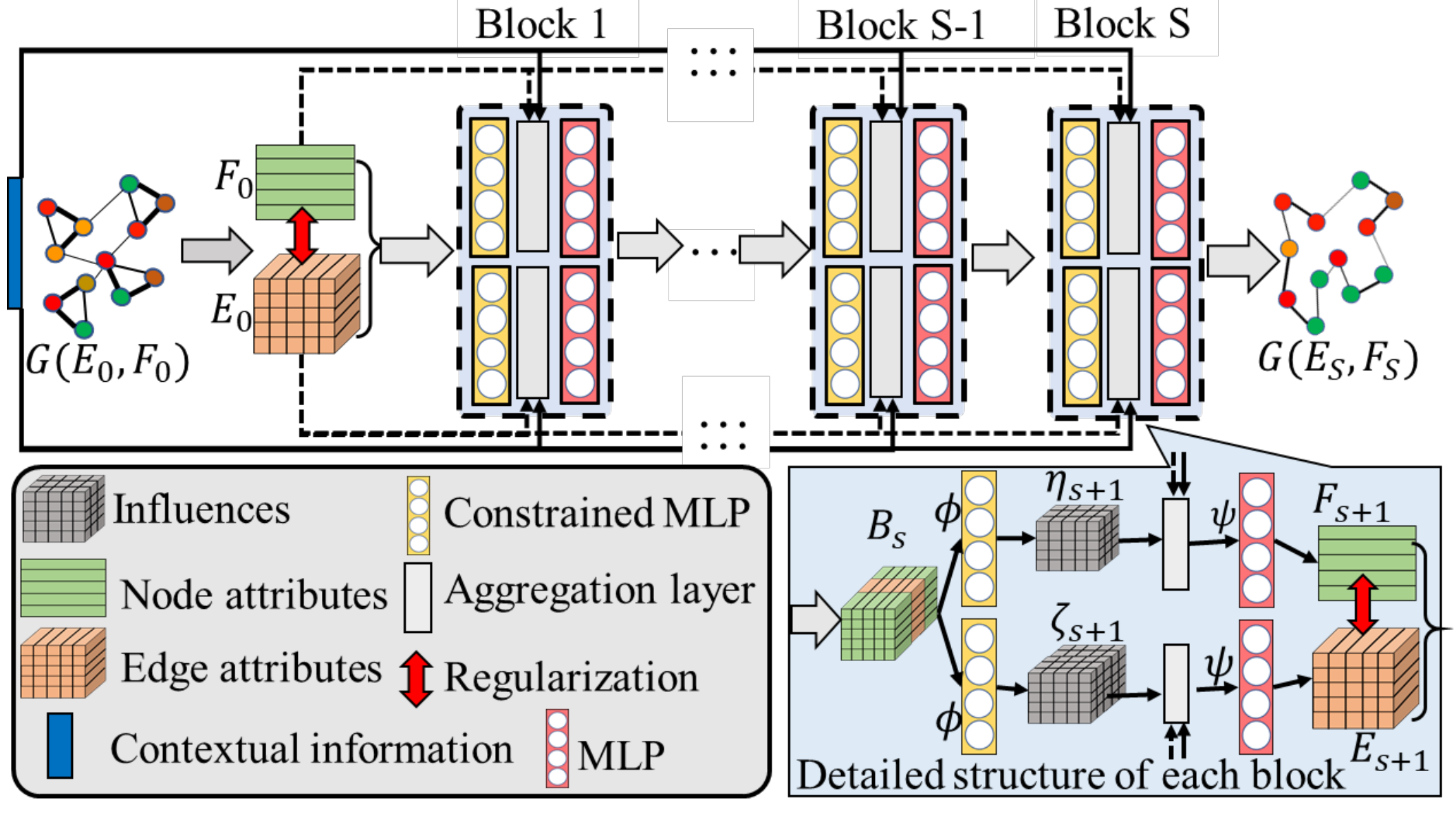}
\caption{The proposed NEC-DGT consists of multiple blocks. Each block has edge and node translation paths which are co-evolved and combined by a graph regularization during training process.}
\label{fig:architecture}
\end{figure}

\textbf{Multi-block asynchronous translation architecture}. The proposed NEC-DGT learns the distribution of graphs in the target domain conditioning on the input graphs and contextual information. However, such a translation process from input graph to the final target graph may experience a series of interactions of different types among edges and nodes. Also, such a sophisticated process is hidden and needs to be learned by a sufficiently flexible and powerful model. To address this, we propose the NEC-DGT as shown in Fig.~\ref{fig:architecture}. Specifically, the node and edge attributes of input graphs are inputted into the model and the model output the generated target graphs' node attributes and edge attributes after several blocks. The skip-connection architecture (black dotted lines in Fig.~\ref{fig:architecture}) implemented across different blocks aims to deal with the asynchrony property of different blocks, which ensures that the final translated results fully utilize various combinations of blocks' information. To train the deep neural network to generate the target graph $G(E',F')$ conditioning on the input graph $G(E_0,F_0)$ and contextual information $C$, we minimize the loss function as follows:
\begin{align}\small
    \mathcal{L}_{\mathcal{T}}=\mathcal{L}( \mathcal{T}(G(E_0,F_0),C), G(E',F'))
\end{align}
where the nodes set $\mathcal{V}_0$ and $\mathcal{V}'$ as well as edges set $\mathcal{E}_0$ and $\mathcal{E}'$ can be reflected in $F_0$ and $F'$, as well as $E_0$ and $E'$.

\textbf{Node and edge translation paths}. To jointly tackle various interactions among nodes and edges, respective translation paths are proposed for each block. In node translation path (in upper part of detailed structure in Fig. \ref{fig:architecture}), node attributes are generated considering the "nodes-to-nodes" and "edges-to-nodes" interactions. In edge translation path (in lower part of detailed structure in Fig. \ref{fig:architecture}), edge attributes are generated following the "edges-to-edges" and "node-to-edges" interactions. 

\textbf{Spectral graph regularization}. To discover and characterize the inherent relationship between nodes and edges of each graph, the frequency domain properties of the graph is learned, based on which the interactions between node and edge attributes are jointly regularized upon non-parametric graph Laplacian. Moreover, to maintain consistent spectral properties throughout the translation process, we enforce the shared patterns among the generated nodes and edges in different blocks by regularizing their relevant parameters in the frequency domain. The regularization of the graphs is formalized as follows:
\begin{align}\small
    \mathcal{R}(G(E,F))=\sum\nolimits_{s=0}^{S} \mathcal{R}_\theta(G(E_s,F_s))+ \mathcal{R}(\theta)
\end{align}
where $S$ refers to the number of blocks, and $\theta$ refers to the overall parameters in the spectral graph regularization. $E_s$ and $F_s$ refer to the generated edge attributes tensor and node attributes matrix in the $s$th block. Thus $G(E_{S},F_{S})$ is the generated target graph. Then the final loss function can be summarized as follows:
\begin{align}\small
    \widetilde{\mathcal{L}}= \mathcal{L}( \mathcal{T}(G(E_0,F_0),C), G(E',F'))+ \beta\mathcal{R}(G(E,F))
\end{align}
where $\beta$ is the trade-off between the $ \mathcal{L}_{\mathcal{T}}$ and spectral graph regularization. The model is trained by minimizing the mean squared error of $E_{S}$ with $E'$, and $F_{S}$ with $F'$, enforced by the regularization. Optimization methods (e.g. Stochastic gradient descent (SGD) and Adam) based on Back-propagation technique can be utilized to optimize the whole model.

Subsequently, the details of a single translation block are introduced: edge translation path in Section~\ref{section: edge}, node translation path in Section~\ref{section: node} and graph spectral-based regularization in Section~\ref{section: graph laplacian}.

\subsection{Edge Translation Path}
\label{section: edge}
\begin{figure}[htb]
\centering
\includegraphics[width=8cm]{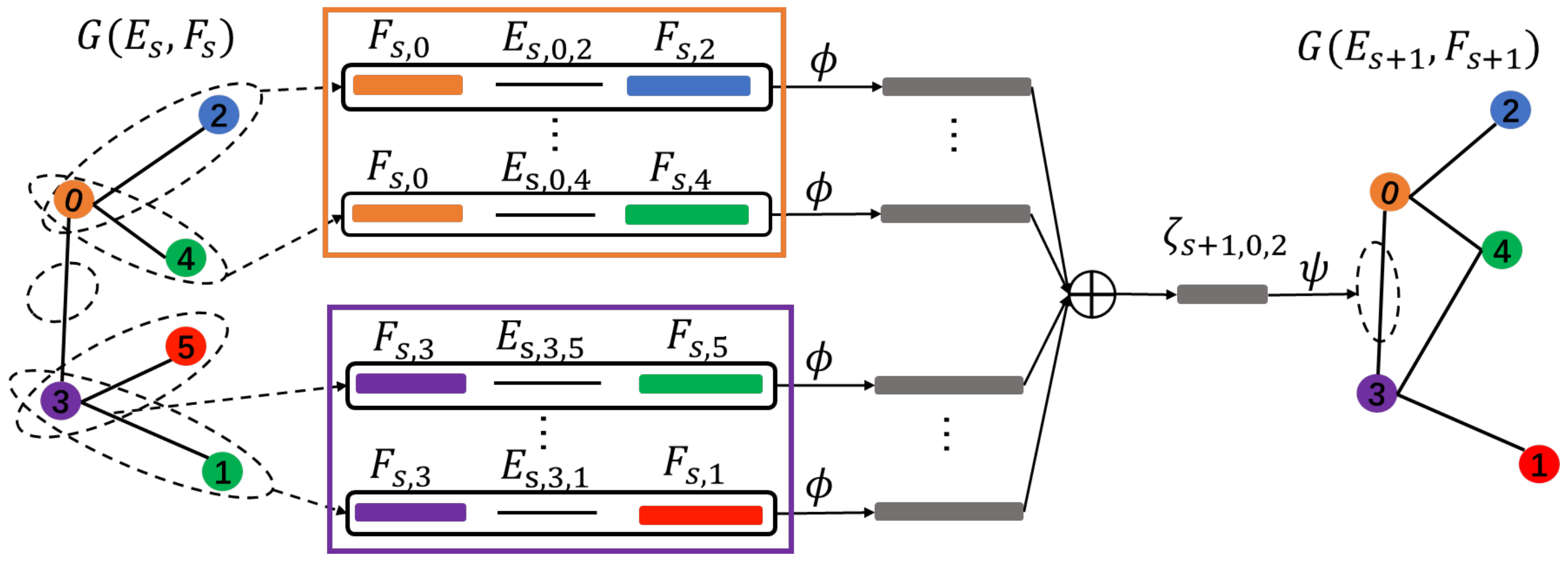}
\caption{Details of edge translation path for one edge (i.e. $e_{0,3}$) in a single block.}
\label{fig:edge path}
\end{figure}

Edge translation path aims to model the nodes-to-edges and edges-to-edges interactions, where edge attributes in the target domain can be influenced by both nodes and edges in the input domain. 
%The existing topology translation methods generate the edge attributes only considering input edge attributes. However, in the multi-attributed graph translation process, both the node and edge attributes influence the translation of edge attributes. 
Therefore, we propose to first jointly embed both node and edge information into influence vectors and then decode it to generate edges attributes. Specifically, the edge translation path of each block contains two functions, \emph{influence-on-edge function} which encodes each pair of edge and node attributes into the influence for generating edges, and the \emph{edge updating function} which aggregates all the influences related to each edge into an integrated influence and decodes this integrated influence to generate each edge' attributes. Fig. \ref{fig:edge path} shows the operation of the two functions in a single block by translating the current input of graph $G(E_s,F_s)$ to output graph $G(E_{s+1},F_{s+1})$.

\textbf{Influence-on-edge layers}. As shown in Fig.~\ref{fig:edge path}, the input graph $G(E_s,F_s)$ is first organized in unit of several pairs of node and edge attributes. For each pair of nodes $v$ and $u$, we concatenate their edge attributes $E_{s,u,v}$ and their node attributes: $F_{s,u}$ and $F_{s,v}$ as: $B_{s,u,v}=\lbrack F_{s,u}, E_{s,u,v}, F_{s,v}\rbrack$ (as circled in black rectangles in Fig.~\ref{fig:edge path}). Then $B_{s,u,v}\in \mathbb{R}^{1\times(2D+K)}$ is inputted into the influence-on-edge function: a constrained MLP (Multilayer Perceptron) $\phi$ which is used to calculate the influence $\phi(B_{s,u,v}) \in \mathbb{R}^{1\times q}$ from the pair of the nodes $u$ and $v$. $q$ refers to the dimension of the final influence on edges. $\phi$ for edge translation path is expressed as follows:

\begin{align}\small
\begin{split}
\phi(X; W_E,b_E)=&\sigma_M(...(\sigma_0(X \cdot W^{(0)}_E+b^{(0)}_E)...W^{(M)}_E+b^{(M)}_E)\\
&s.t., W^{(0)}_{E,1:D}\equiv W^{(0)}_{E,(D+K):(2D+K)} 
\end{split}
\label{mlp}
\end{align}

where $W_E$ and $b_E$ are weights and bias for $\phi$ in edge translation path. $M$ refers to the number of layers of $\phi$ and $\{\sigma_0,...\sigma_M\}$ refers to the activation functions. For undirected graph, we add a weight constraint $\scriptsize {W^{(0)}_{E,1:D}\equiv W^{(0)}_{E,(D+K):(2D+K)}}$ to ensure that the influence of $B_{s,u,v}$ is the same as the influence of $B_{s,v,u}$, which means that the first $D$ rows (related to the attributes of node $u$ ) and the last $D$ rows (related to the attributes of node $v$) of $W^{(0)}_E$ are shared. The influence on edges of each pair is computed through the same function with the same weights. Thus the NEC-DGT can handle various size of graphs. 

\textbf{Edge updating layers}. After calculating the influence of each pair of nodes and edge, the next step is to assigning each pairs' influences to its related edge to get the integrated influence for each edge~(as shown of $\bigoplus$ operation in Fig.\ref{fig:edge path}). This is because each edge is generated depending on both its two related nodes and its incident edges (like the pairs circled in the orange rectangle and purple rectangle related to node $0$ and node $3$ respectively in Fig.\ref{fig:edge path}). Here we define the integrated influence on one edge attribute $E_{s+1,i,j}$ as: $\zeta_{s+1,i,j} \in \mathbb{R}^{1\times q}$, which is computed as follows:

\begin{align}\small
\begin{split}
  {\zeta}_{s+1,i,j}=\!\sum\nolimits_{k_1\in N(i)}\!\phi(B_{s,i,k_1};W_E,b_E) +\\
  \!\sum\nolimits_{k_2\in N(j)}\!\phi(B_{s,k_2,j};W_E,b_E) 
  \label{eq:edge sum}
  \end{split}
\end{align}

where $N(i)$ refers to the neighbor nodes of node $i$. Then the edge attributes $E_{s+1,i,j}$ is generated by $\psi([E_{0,i,j}, \zeta_{s+1,i,j},C])$, where $E_{0,i,j}$ refers to the input edge attributes of edge $e_{i,j}$. $C$ refers to the contextual information for the translation. The function $\psi$ is implemented by an MLP.

\textbf{Relationship with other edge convolution networks}.
Edge convolution network is the most typical method to handle the edge embedding in graphs, which was first introduced as BrainNetCNN~\cite{kawahara2017brainnetcnn} and later explored in many studies~\cite{guo2018deep,kivilcim2018modeling,sturmfels2018domain}. Our edge translation path is a highly flexible and generic mechanism to handle multi-attributed nodes and edges. Several existing edge convolution layers and their variants can be considered as special cases of our method, as demonstrated in the following theorem{\small \footnote{The proof process is available at\url{https://github.com/xguo7/NEC-DGT}}}:

\begin{theorem}
\label{theorem}
The influence-on-edge function $\phi$ in edge translation path of NEC-DGT is a generalization of conventional edge convolution networks.
\end{theorem}

\subsection{Node Translation Path}
\label{section: node}
\begin{figure}[htb]
\centering
\includegraphics[width=9cm]{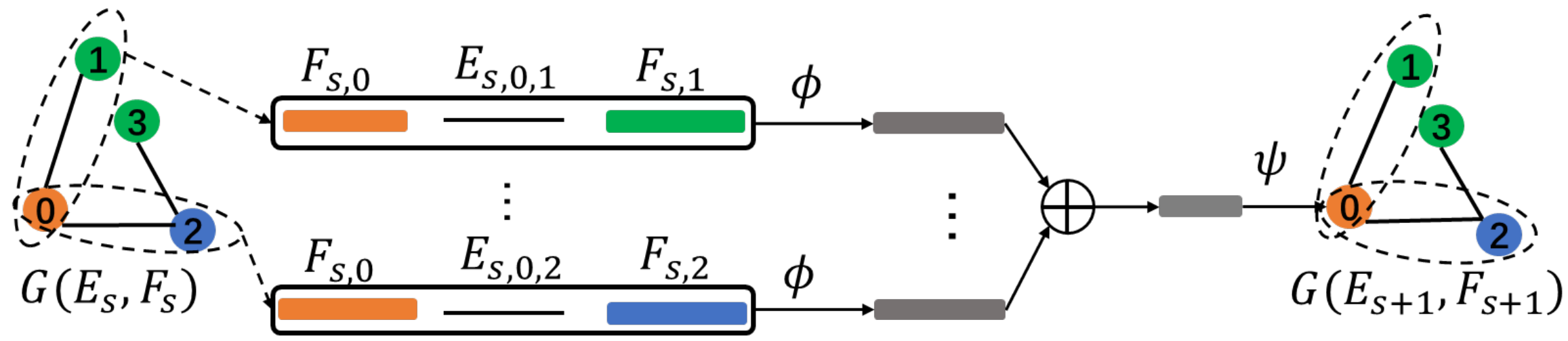}
\caption{Details of node translation path for one node (i.e. node 0) in a single block.}
\label{fig:node path}
\end{figure}

Node translation aims to learn the ``nodes-to-nodes" and ``edges-to-nodes" interactions, where translation of one node's attributes depends on the edge attributes related to this node and its own attributes. The node translation path of each block contains two functions, \emph{influence-on-node function} which learns the influence from each pair of nodes, and \emph{node updating function} which generates the new node attributes by aggregating all the influences from pairs containing this node. Fig. \ref{fig:node path} shows how to translate a node in a single block.

\textbf{Influence-on-node layers}. As shown in Fig. \ref{fig:node path}, the input graph $G(E_s, F_s)$ is first organized in the unit of pairs of nodes, where each pair is $B_{s,u,v} \in  \mathbb{R}^{1 \times (2D+K)}$ which is similar to the edge translation path (as circled in the black rectangle in Fig.~\ref{fig:node path}). Then $B_{s,u,v}$ is inputted into the influence-on-node function, which is implemented by contrained MLP $\phi$ as Equation \eqref{mlp}, to compute the influence $\phi(B_{s,u,v};W_F,b_F)\in \mathbb{R}^{1\times h}$  to nodes (as shown in the grey bar after $\phi$ in Fig.~\ref{fig:node path}), where $h$ is the dimension of the influence on nodes.

\textbf{Node updating layers}. After computing the influences of each node pair, the next step is to generate node attributes. For node $i$, an assignment step is required to aggregate all the influences from pairs containing node $i$ (as shown of $\bigoplus$ operation in Fig.~\ref{fig:node path}). Thus, all the influences for node $i$ are aggregated and input into the updating function, which is implemented by a MLP model $\psi$ to calculate the attributes of node $i$ as: $ F_{s+1,i}= \psi([F_{0,i},\sum_{j\in N(i)}\phi(B_{s,i,j};W_F,b_F),C])$. 

\subsection{Graph spectral-based regularization}
\label{section: graph laplacian}
Based on the edge and node translation path introduced above, we can generate node and edge attributes, respectively. However, since these generated node and edge attributes are predicted separately in different paths, their patterns may not be consistent and harmonic. To ensure the consistency of the edge and node patterns mentioned in Section \ref{section: problem}, we propose a novel adaptive regularization based on non-parametric graph Laplacian, and a graph frequency regularization.

\textbf{Non-parametric Graph Laplacian Regularization}.
First, we recall the property of the multi-attributed graphs where node information can be smoothed over the graph via some form of explicit graph-based regularization, namely, by the well-known graph Laplacian regularization term~\cite{kipf2016semi}: 
\begin{small}
${F_{s}^{(d)}}^T\!L_{s}^{(k)} F_{s}^{(d)}=\!\sum_{i,j\in\mathcal{V}}E_{s,i,j}^{(k)}\left\|F_{s,i}^{(d)}-F_{s,j}^{(d)}\right\|^2$,
\end{small}
where $F_{s}^{(d)}\!\in\mathbb{R}^{N\times 1}$ is the node attribute vector for the $d$th node attribute and $E_{s}^{(k)}\!\in\mathbb{R}^{N\times N}$ is the edge attribute matrix for $k$th attribute generated in the $s$th block. $L_s^{(k)}\!=\!D_s^{(k)}-E_s^{(k)}$ denotes the graph Laplacian for the $k$th edge attributes matrix. The degree matrix $D_s^{(k)}\in \mathbb{R}^{N\times N}$ is computed as: $D_{s,i,i}^{(k)} =\sum_{j \in N(i)} E_{s,i,j}^{(k)}$.

However, the above traditional graph Laplacian can only impose an absolute smoothness regularization over all the nodes by forcing the neighbor nodes to have similar attribute values, which is often over-restrictive for many situations such as in signed networks and teleconnections. In the real world, the correlation among the nodes is much more complicated than purely "smoothness" but should be a mixed pattern of different types of relations. To address this, we propose an end-to-end framework of non-parametric graph Laplacian which can automatically learn such node correlation patterns inherent in specific types of graphs, with rigorous foundations on spectral graph theory. In essence, we propose the non-parametric graph Laplacian based on the parameter $\theta$ as: $g_\theta(\hat{L}_s^{(k)})$. $\hat{L}_s^{(k)}$ is the normalized Laplacian computed as
\begin{small}
$\small{\hat{L}_s^{(k)}={D_s^{(k)}}^{-\frac{1}{2}}L_s^{(k)}{D_s^{(k)}}^{-\frac{1}{2}}}$ 
\end{small}
and can be diagonalized by the Fourier basis $U_s^{(k)} \in \mathbb{R}^{N \times N}$, such that $\hat{L}_s^{(k)} = U_s^{(k)}\Lambda_s^{(k)} {U_s^{(k)}}^T$ where $\Lambda_s^{(k)}\!\in\! \mathbb{R}^{N \times N}$ is a diagonal matrix storing the graph frequencies. For example, $\Lambda_{s,1}^{(k)}$ is the frequency value of the first Fourier basis $U_{s,1}^{(k)}$. Then we got $g_\theta(\hat{L}_s^{(k)})=g_\theta(U_s^{(k)}\Lambda_s^{(k)} {U_s^{(k)}}^T)=U_s^{(k)} g_\theta(\Lambda_s^{(k)}){U_s^{(k)}}^T$. 
Therefore, we have the regularization as follows:

\begin{align}\small\nonumber
\begin{split}
    \mathcal{R}_\theta (G(E_s,F_s))=\sum\nolimits_{k=1}^{K}\sum\nolimits_{d=1}^{D}{F_s^{(d)}}^T U_s^{(k)} g_\theta(\Lambda_s^{(k)}){U_s^{(k)}}^T F_s^{(d)}
    \end{split}
\end{align}

where $g_\theta(\Lambda_s^{(k)})$ is a non-paramteric Laplacian eigenvalues that will be introduced subsequently.

\textbf{Scalable approximation}.
$g_\theta(\Lambda_s^{(k)})$ is a non-parametric vector whose parameters are all free; It can be defined as: $g_\theta(\Lambda_s^{(k)})=diag(\theta_s^{(k)})$, where the parameter $\theta_s^{(k)} \in \mathbb{R}^N$ is a vector of Fourier coefficients for a graph.
However, optimizing the parametric eigenvalues has the learning complexity of $O(N)$, the dimensionality of the graphs, which is not scalable for large graphs. To reduce the learning complexity of $O(N)$ to $O(\textbf{1})$, we propose approximating $g_\theta(\Lambda_s^{(k)})$ by a normalized truncated expansion in terms of Chebyshev polynomials~\cite{hammond2011wavelets}. The Chebyshev polynomial $T_p(x)$ of order p may be computed by the stable recurrence relation $T_p(x)=2xT_{p-1}(x)-T_{p-2}(x)$ with $T_1 = 1$ and $T_2 = x$. The eigenvalues of the approximated Laplacian filter can thus be parametric as the truncated expansion:
\begin{align}\small
\begin{split}
   g_\theta(\Lambda_s^{(k)})=\sum\nolimits^{P}_{p=1}\theta_{s,p}^{(k)} T_p(\widetilde{\Lambda}_s^{(k)})/\sum\nolimits^{P}_{p=1}\theta_{s,p}^{(k)}
   \end{split}
\end{align}

for $P$ orders, where $T_p(\widetilde{\Lambda}_s^{(k)}) \in \mathbb{R}^{N \times N}$is the Chebyshev polynomial of order $p$ evaluated at $\widetilde{\Lambda}_s^{(k)}=2\Lambda_s^{(k)}/\Lambda_{s,max}^{(k)}-I $, a diagonal matrix of scaled eigenvalues that lie in $[-1, 1]$. The $\Lambda_{s,max}$ refers to the largest element in $\Lambda_s^{(k)}$. $\theta \in \mathbb{R}^{S \times P \times K}$ denotes the parameter tensor for all $S$ blocks. $\theta_{s,p}^{(k)}$ is the $p$th element of Chebyshev coefficients vector $\theta_{s}^{(k)}\in \mathbb{R}^P$ for the $k$th edge attribute. Each $\theta_{s,p}^{(k)}$ is normalized by dividing the sum of all the coefficients in $\theta_s^{(k)}$ to avoid the situation where $\theta_s^{(k)}$ is trained as zero. Thus, the laplacian computation can then be written as $g_\theta(\widetilde{L}_s^{(k)})=\sum^{P}_{p=1}\theta_{s,p}^{(k)}T_p(\widetilde{L}_s^{(k)}) / \sum^{P}_{p=1}\theta_{s,p}^{(k)} $, where $T_p(\widetilde{L}_s^{(k)}) \in \mathbb{R}^{N \times N}$ is the Chebyshev polynomial of order $p$ evaluated at the scaled Laplacian $\widetilde{L}_s^{(k)}=2\hat{L}_s^{(k)}/\Lambda_{s,max}^{(k)}-I $. For efficient computation, we further approximate $\Lambda_{s,max}^{(k)}\thickapprox 1.5$, as we can expect that the neural network parameters $\theta$ will adapt to this change in scale during training. 

\textbf{Graph frequency regularization}.
To ensure that the spectral graph patterns are consistent throughout the translation process across different blocks, we utilize a graph frequency regularization to not only maintain the similarity but also allow the exclusive properties of each block's patterns to be reserved to some degree. Specifically, regarding all the frequency pattern basis of form $\widetilde{L}$, some are important in modeling the relationships between nodes and graphs while some are not, resulting in the sparsity pattern of $\theta$. Thus, inspired by the multi-task learning, we learn the consistent sparsity pattern of $\theta_s$ by using the $L_{2,1}$ norm as regularization:
\begin{align}\small
\begin{split}
    \mathcal{R}(\theta)=\sum\nolimits_{k=1}^{K}\sum\nolimits_{s=1}^{S} \sqrt{\sum\nolimits_{p=1}^{P}{\theta_{s,p}^{(k)}}^2}
    \end{split}
\end{align}

\subsection{Complexity Analysis}
The proposed NEC-DGT requires $O(N^2)$ operations in time complexity and $O(N^2)$ space complexity in terms of number of nodes in the graph. It is more scalable than most of the graph generation methods. For example, GraphVAE~\cite{simonovsky2018graphvae} requires $O(N^4)$ operations in the worst case and Li et al~\cite{li2018learning} uses graph neural networks to perform a form of “message passing” with $O(MN^2)$ operations to generate a graph.

\section{Experiments}
In this section, we present both the quantitative and qualitative experiment results on NEC-DGT as well as the comparison models. All experiments are conducted on a 64-bit machine with Nvidia GPU (GTX 1070, 1683 MHz, 8 GB GDDR5). The model is trained by ADAM optimization algorithm{\small \footnote{The code of the model and additional experiment results are available at:\url{https://github.com/xguo7/NEC-DGT}}}.

\subsection{Experimental Setup}
\subsubsection{Datasets} We performed experiments on four synthetic datasets and four real-world datasets with different graph sizes and characteristics. All the dataset contain input-target pairs.

\textbf{Synthetic dataset}: 
Four datsets are generated based on different types of graphs and translation rules. The input graphs of the first three datasets (named as Syn-I, Syn-II, and Syn-III) are Erdos-Renyi (E-R) graphs generated by the Erdos Renyi model~\cite{erdHos1960evolution} with the edge probability of 0.2 and graph size of 20, 40, and 60 respectively. The target graph topology is the 2-hop connection of the input graph, where each edge in the target graph refers to the 2-hop reachability in the input graph (e.g. if node $i$ is 2-hop reachable to node $j$ in the input graph, then they are connected in the target graph). The input graphs of the fourth dataset (named as Syn-IV) are Barab\'asi-Albert (B-A) graphs generated by the Barab\'asi-Albert model~\cite{barabasi1999emergence} with 20 nodes, where each node is connected to 1 existing node. In Syn-IV, topology of target graph is the 3-hop connection of the input graph. For all the four datasets, the edge attributes $E_{s,i,j} \in[0,1]$ denotes the existence of the edge. For both input and target graphs, the node attributes are continuous values computed following the polynomial function: $f(x):y=ax^2+bx+c (a=0,b=1,c=5)$, where $x$ is the node degree and $f(x)$ is the node attribute.  Each dataset is divided into two subsets, each of which has 250 pairs of graphs. Validation is conducted where one subset is used for training and another for testing, and then exchange them for another validation. The average result of the two validations is regarded as the final result.

\textbf{Malware confinement dataset}:
Malware dataset are used for measuring the performance of NEC-DGT for malware confinement prediction. There are three sets of IoT nodes at different amount (20, 40 and 60) encompassing temperature sensors connected with Intel ATLASEDGE Board and Beagle Boards (BeagleBone Blue), communicating via Bluetooth protocol. Benign and malware activities are executed on these devices to generate the initial attacked networks as the input graphs. Benign activities include MiBench~\cite{guthaus2001mibench} and SPEC2006~\cite{henning2006spec}, Linux system programs, and word processor. The nodes represent devices and node attribute is a binary value referring to whether the device is compromised or not. Edge represents the connection of two devices and the edge attribute is a continuous value reflecting the distance of two devices. The real target graphs are generated by the classical malware confinement methods: stochastic controlling with malware detection~\cite{sayadi2018ensemble,DATE'19_IoT,DATE'19_2smart}. We collected 334 pairs of input and target graphs with different contextual parameters (infection rate, recovery rate, and decay rate) for each of the three datasets. Each dataset is divided into two subsets: one has 200 pairs and another has 134 pairs. The validation is conducted in the same way as the synthetic dataset.

\textbf{Molecule reaction dataset}:
We apply our NEC-DGT to one of the fundamental problems in organic chemistry, thus predicting the product (target graph) of chemical reaction given the reactant (input graph). Each molecular graph consists of atoms as nodes and bond as edges. The input molecule graph has multiple connected components
since there are multiple molecules comprising the reactants. The reactions used for training are atom-mapped so that each atom in the product graph has a unique corresponding atom in the reactants. We used reactions from USPTO granted patents, collected by Lowe~\cite{lowe2014patent}. we obtained a set of 5,000 reactions (reactant-product pair) and divided them into 2,500 and 2,500 for training and testing. Atom (node)  features include its elemental identity, degree of connectivity, number of attached hydrogen atoms, implicit valence, and aromaticity. Bond (edge) features include bond type (single, double, triple, or aromatic), and whether it is connected.

\subsubsection{Comparison methods} Since there is no existing method handling the multi-attributed graph translation problem, NEC-DGT is compared with two categories of methods: 1) graph topology generation methods, and 2) graph node attributes prediction methods.

\textbf{Graph topology generation methods}:
1) GraphRNN~\cite{you2018graphrnn} is a recent graph generation method based on sequential generation with LSTM model; 2) Graph Variational Auto-encoder (GraphVAE)~\cite{simonovsky2018graphvae} is a VAE based graph generation method for small graphs; 3) Graph Translation-Generative Adversarial Networks (GT-GAN)~\cite{guo2018deep} is a new graph topology translation method based on graph generative adversarial network.

\textbf{Node attributes prediction methods}:
1) Interaction Network (IN)~\cite{battaglia2016interaction} is a node state updating network considering the interaction of neighboring nodes; 2) DCRNN~\cite{li2017diffusion} is a node attribute prediction network for tranffic flow prediction; 3) Spatio-Temporal Graph Convolutional Networks (STGCN)~\cite{yu2017spatio} is a node attribute prediction model for traffic speed forecast.

Furthermore, to validate the effectiveness of the graph spectral-based regularization, we conduct a comparison model (named as NR-DGT) which has the same architecture of NEC-DGT but without the graph regularization.

\subsubsection{Evaluation metrics}
A set of metrics are used to measure the similarity between the generated and real target graphs in terms of node and edge attributes. To measure the attributes which are Boolean values, the Acc (accuracy) is utilized to evaluate the ratio of nodes or edges that are correctly predicted among all the nodes or possible node pairs. To measure the attributes which are continuous values, MSE (mean squared error), R2 (coefficient of determination score), Pearson and Spearman correlation are computed between attributes of generated and real target graphs. $N-<metric>$ represents metrics evaluated on node attributes and $E-<metric>$ represents metrics evaluated on edge attributes.

\subsection{Performance}
\subsubsection{Metric-based evaluation for synthetic graphs} For synthetic datasets, we compare the generated and real target graphs on various metrics and visualize the patterns captured in the generated graphs.
%\paragraph{\textbf{Metric-based evaluation}}\vspace{-0.2cm}
Table \ref{table:synthetic dataset I} summarizes the effectiveness comparison for four synthetic datasets. The node attributes are continuous values evaluated by N-MSE, N-R2, N-P, and N-SP. The edge attributes are binary values evaluated by the accuracy of the correctly predicted edges. 
The results in Table \ref{table:synthetic dataset I} demonstrate that the proposed NEC-DGT outperforms other methods in both node and edge attributes prediction and is the only method to handle both. Specifically, in terms of node attributes, the proposed NEC-DGT get smaller N-MSE value than all the node attributes prediction methods by 85\%, 71\%, 95\% and 95\% on average for four dataset respectively. Also, NEC-DGT outperforms the other methods by 46\%, 36\%, 44\% and 58\% on average for four dataset respectively on N-R2, N-P, and N-SP. This is because all the node prediction methods only consider a fixed graph topology while NEC-DGT allows the edges to vary. In terms of edges, the proposed NEC-DGT get the highest E-ACC than all the other graph generation methods. It also has higher E-ACC than graph topology translation method: GT-GAN by 7\% on average since NEC-DGT considers both edge and node attributes in learning the translation mapping while GT-GAN only considers edges. The proposed NEC-DGT outperforms the NR-DTG by around 3\% on average in terms of all metrics, which demonstrates the effectiveness of the graph spectral-based regularization.

\begin{table}[htb]\scriptsize\renewcommand\arraystretch{1.1}
  \centering
   \caption{Evaluation of Generated Target Graphs for Synthetic Dataset (N for node attributes, E for edge attributes, P for Pearson correlation, SP for Spearman correlation and Acc for accuracy).\vspace{-0.3cm}}
  \begin{tabular}{p{0.55cm}p{1.08cm}rp{0.6cm}rr|p{1.08cm}p{0.65cm}}           \\
   \toprule dataset & Method   & N-MSE   &N-R2    &N-P &N-Sp& Method&E-Acc  \\
    \hline 
    \multirow{8}{*}{Syn-I} 
    %&GraphRNN &- &- &- &- &62.12\% \\
     %~ &GraphVAE &- &- &- &- &65.91\% \\
    %~ &GT-GAN &- &- &- &- &70.39\% \\
    &IN &5.97 &0.06 &0.48 &0.44	&GraphRNN &0.6212\\
    ~ &DCRNN &51.36	&0.12	&0.44	&0.45	&GraphVAE&0.6591\\
    ~ &STGCN &15.44	&0.19	&0.42	&0.56	&GT-GAN&0.7039\\
    ~ &NR-DGT &2.13	&\textbf{0.87}	&0.90	&0.89&NR-DGT&0.7017\\
    ~ &NEC-DGT &\textbf{1.98} &0.76 &\textbf{0.93} &\textbf{0.91}&NEC-DGT &\textbf{0.7129}\\
    \toprule   
    \multirow{8}{*}{Syn-II} 
    %&GraphRNN &- &- &- &- &56.21\% \\
     %~ &GraphVAE &- &- &- &- &46.39\% \\
    %~ &GT-GAN &- &- &- &- &70.05\%\\
    &IN &\textbf{1.36}	&0.85	&0.77	&0.87 &GraphRNN&0.5621\\
    ~&DCRNN &71.07	&0.11	&0.39	&0.37 &GraphVAE&0.4639\\
    ~ &STGCN &33.11	&0.21	&0.15	&0.15&GT-GAN &0.7005\\
    ~ &NR-DGT &1.43 &0.91 &0.94 &\textbf{0.97}&NR-DGT &0.7016\\
    ~ &NEC-DGT &1.91 &\textbf{0.93} &\textbf{0.97} &\textbf{0.97}&NEC-DGT &\textbf{0.7203}\\
  \toprule
    \multirow{8}{*}{Syn-III}%& -& -&- &- \\
     %~  & -& -&- &-  \\
    %~ & &- &-  \\
    &IN &35.46 &0.31 &0.59 &0.56&GraphRNN &0.4528\\
    ~ &DCRNN &263.23 &0.09 &0.41 &0.39 &GraphVAE&0.3702\\
    ~ &STGCN & 43.34&0.22 &0.48 &0.47 &GT-GAN&0.5770\\
    ~ &NR-DGT &5.90 &0.90 &0.94 &0.92 &NR-DGT&0.6259\\
    ~ &NEC-DGT &\textbf{4.56} &\textbf{0.93} &\textbf{0.97} &\textbf{0.96} &NEC-DGT&\textbf{0.6588}\\
    \toprule
    \multirow{8}{*}{Syn-IV}
    % & -& -&- &-  \\
     %~  & -& -&-   \\
    %~  &- &- &- \\
     &IN &4.63 &0.10 &0.53 &0.51&GraphRNN&0.5172\\
    ~ &DCRNN &63.03 &0.12 &0.22 &0.16 &GraphVAE&0.3001\\
    ~ &STGCN & 6.52&0.08 &0.11 &0.10 &GT-GAN&0.8052\\
    ~ &NR-DGT &4.49 &0.12 &0.55 &0.54 &NR-DGT&0.6704\\
    ~ &NEC-DGT &\textbf{1.86} &\textbf{0.73} &\textbf{0.93} &\textbf{0.89} &NEC-DGT&\textbf{0.8437}\\
    \bottomrule
  \end{tabular}
  \label{table:synthetic dataset I}
\end{table}

\subsubsection{Evaluation of the learned translation mapping for synthetic graphs}
\begin{figure}[htb]
\centering
\includegraphics[width=8cm]{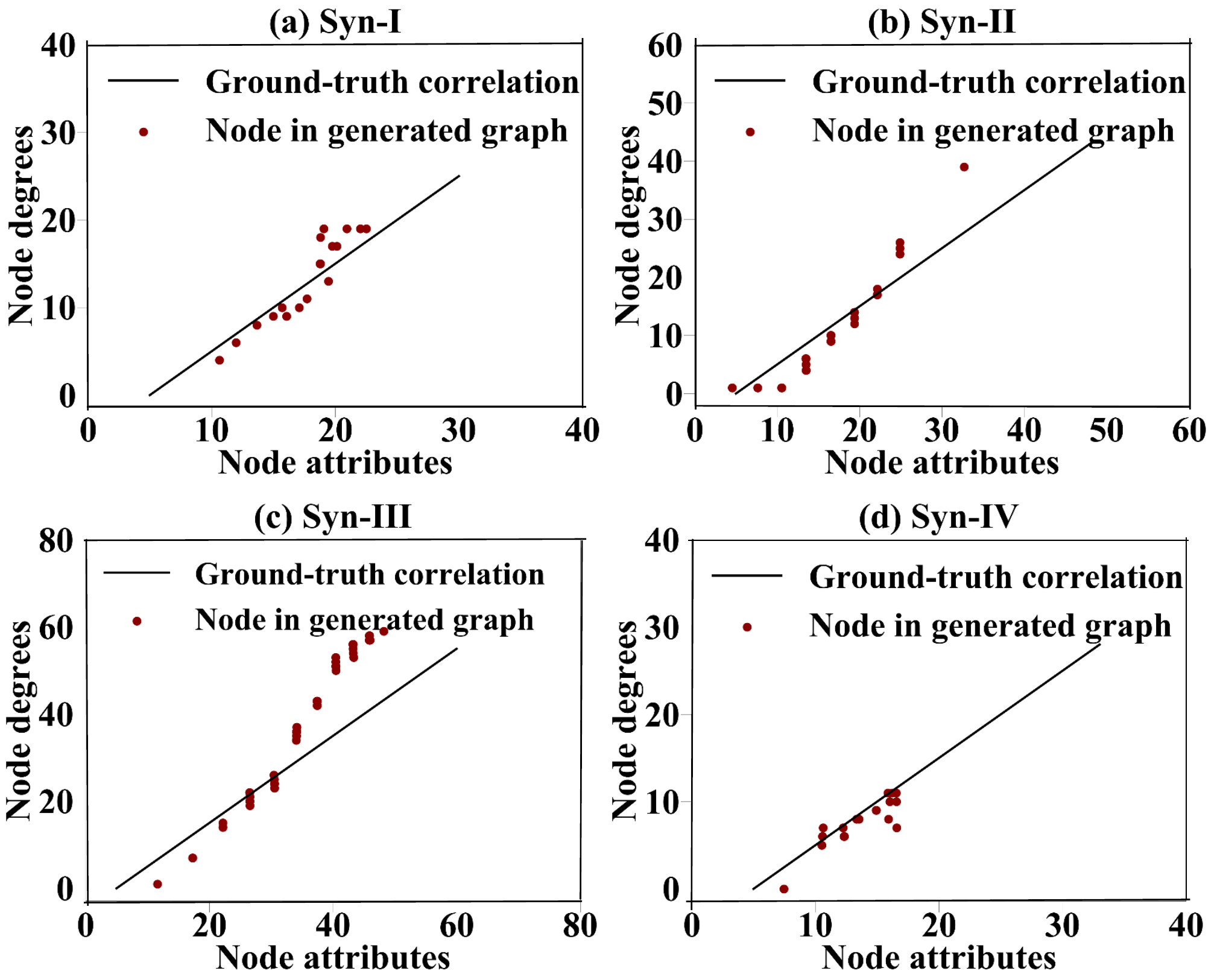}
\caption{Relation visualizations between node attributes and node degrees for samples from four synthetic graphs}
\label{fig:erdos}
\end{figure}

To evaluate whether the inherent relationship between node and edge (reflected by node degree) attributes is learned and maintained by NEC-DGT, we draw the distributions of the node attribute versus node degree of each node in the generated graphs to visualize their relationship. For comparison, a ground-truth correlation is drawn according to the predefined rule of generating the dataset, namely, each node's degree and attribute follows the function $y=x+5$. Fig.~\ref{fig:erdos} shows four example distributions of nodes in terms of node attributes and degree with the black line as ground-truth. As shown in Fig.~\ref{fig:erdos}, the nodes are located closely on the ground-truth, especially for the syn-I and syn-IV, where around 85\% nodes are correctly located. This is largely because the proposed graph spectral-based regularization successfully discovers the patterns: the densely connected nodes all tend to have large node attributes and in reverse. 

\subsubsection{Metric-based Evaluation for malware datasets}
Table \ref{table:malware} shows the evaluation of NEC-DGT by comparing the generated and real target graphs. For malware graphs, the node attributes are evaluated by N-ACC by calculating the percentage of nodes whose attributes are correctly predicted in all nodes. The edge attributes are continuous value evaluated by E-MSE, E-R2 and E-P. We also use E-Acc to evaluate the correct existence of edges among all pairs of nodes. The results in Table \ref{table:malware} demonstrates that NEC-DGT performs the best for all the three datasets. In terms of E-Acc, the graph generation methods (GraphRNN and GraphVAE) cannot handle the graph translation work and got low E-Acc of around 0.6 at Mal-I,Mal-II, and 0.8 at Mal-III. GT-GAN achieves high E-ACC, but its E-MSE is about 2 folds larger than that of the proposed NEC-DGT on average. NEC-DGT successfully handle the translation tasks with high E-Acc above 0.9, and the smallest E-MSE. In terms of N-Acc, NEC-DGT outperforms other methods by around 5\% on the first two datasets. In summary, the proposed NEC-DGT can not only jointly predict the node and edges attributes, but also performs the best in most of metrics. The superiority of NEC-DGT over the NR-DGT in terms of E-MSE demonstrates that the graph spectral-based regularization indeed improve modeling translation mapping.  
\begin{table}[h]\scriptsize\renewcommand\arraystretch{1.1}
  \centering
  %\captionsetup{font={small}}
   \caption{Evaluation of Generated Target Graphs for Malware Dataset (N for node attributes, E for edge attributes, P for Pearson correlation, SP for Spearman correlation and Acc for accuracy).\vspace{-0.3cm}}
  \begin{tabular}{p{1.08cm}p{0.7cm}rp{0.55Cm}p{0.4Cm}|p{1.08cm}p{0.7cm}}\\    
     \toprule 
    \multicolumn{7}{c}{Malware-I}\\
    \hline
    Method&E-Acc& E-MSE   &E-R2    &E-P & Method&N-Acc\\
    \hline 
    GraphRNN &0.6107 &1831.43 &0.52&0.00  &IN&0.8786\\
      GraphVAE &0.5064 &2453.61	&0.00	&0.04 &DCRNN &0.8786\\
    GT-GAN &0.6300 &1718.02	&0.42&0.11 &STGCN&0.9232\\
    %~ &IN &- &- &- &- &- &87.86\% \\
    %~ &DCRNN &- &- &- &- &- &87.86\%\\
    %~ &STGCN &- &- &- &- &- &92.32\%\\
     NR-DGT &0.9107 &668.57 &\textbf{0.82} &\textbf{0.91} &NR-DGT& 0.9108\\
     NEC-DGT &\textbf{0.9218}&\textbf{239.79} &0.78 &\textbf{0.91}&NEC-DGT&\textbf{0.9295}\\
  \toprule
     \multicolumn{7}{c}{Malware-II}\\
     \hline 
      Method&E-Acc& E-MSE   &E-R2    &E-P & Method&N-Acc\\
       \hline 
    GraphRNN &0.7054 & 1950.46& 0.44&0.29 &IN&0.8828\\
     GraphVAE &0.6060 &2410.57 & 0.73&0.16 &DCRNN&0.8790\\
    GT-GAN &0.9033 &462.73	&0.13	&0.81 &STGCN&0.9330\\
    %~ &IN  &- &- &- &- &- &88.28\%\\
    %~ &DCRNN &- &- &- &- &- &87.90\%\\
    %~ &STGCN  &- &- &- &- &- &93.30\%\\
     NR-DGT &0.9117 &448.48 &0.68 &0.83&NR-DGT&0.8853\\
     NEC-DGT &\textbf{0.9380} &\textbf{244.40} &\textbf{0.81} &\textbf{0.91}&NEC-DGT&\textbf{0.9340}\\
  \toprule 
    \multicolumn{7}{c}{Malware-III}\\
     \hline 
    Method&E-Acc& E-MSE   &E-R2    &E-P & Method&N-Acc\\
       \hline 
    GraphRNN &0.8397&1775.58 &0.16 &0.23&IN&0.8738\\      GraphVAE &0.8119 &2109.64 & 0.39&0.32 &DCRNN&0.8738\\
     GT-GAN & 0.9453&550.30 &0.63 &0.80&STGCN&\textbf{0.9375}\\
    %~ &IN & -&- &- &-&-&87.38\%\\
    %~ &DCRNN &- &- &- &-&-&87.38\%\\
    %~ &STGCN &- &- &- &-&-&\textbf{93.75}\%\\
     NR-DGT &0.9543 &341.10 &0.76 &0.88&NR-DGT&0.8773\\
     NEC-DGT&\textbf{0.9604} &\textbf{273.67} &\textbf{0.81} &\textbf{0.90}&NEC-DGT&0.9002\\
\bottomrule
  \end{tabular}
  \label{table:malware}
\end{table}

\subsubsection{Case study for malware dataset}
\begin{figure}[htb]
\centering
\includegraphics[width=8cm]{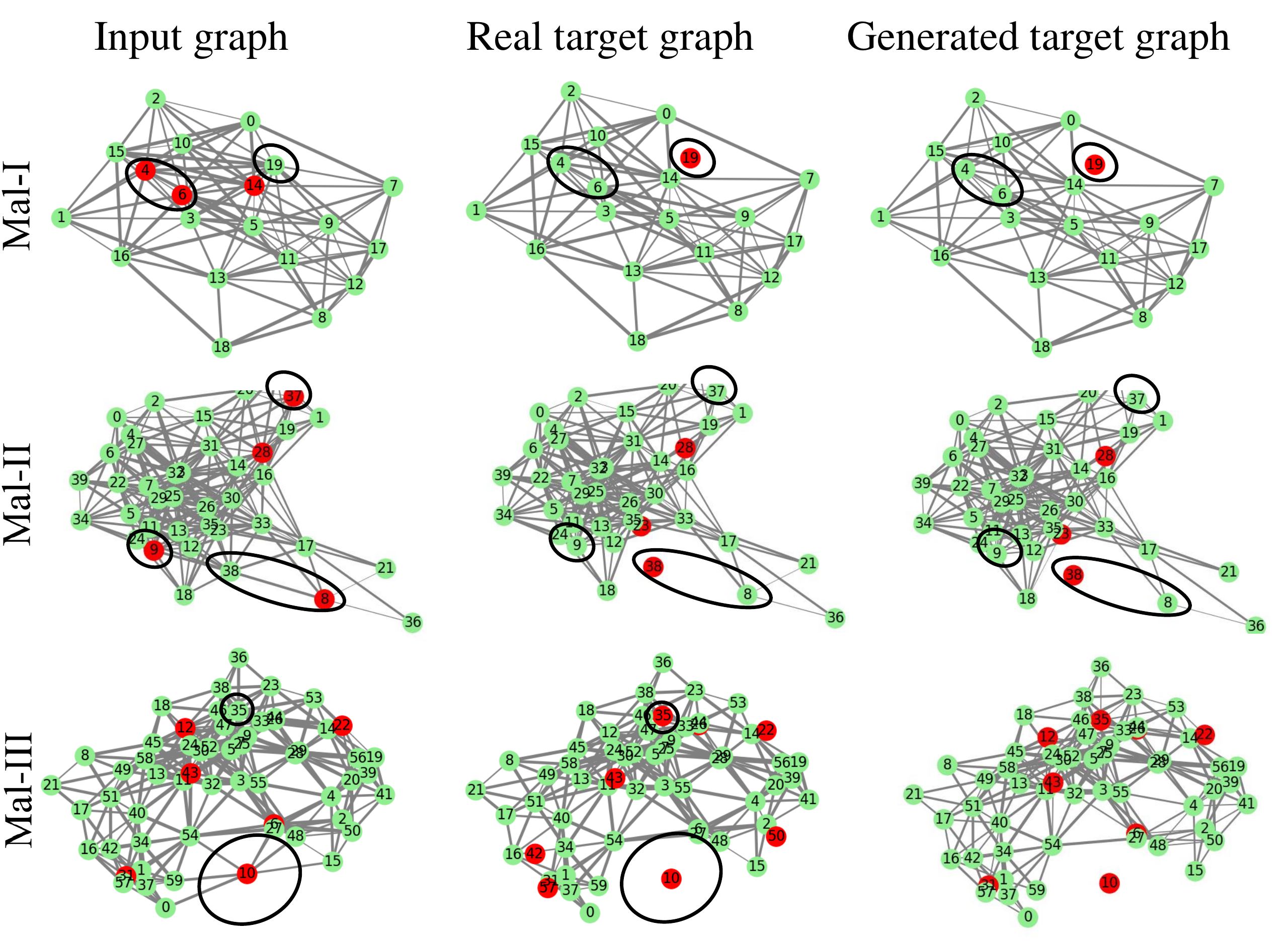}
\caption{Cases of Malware translation by NEC-DGT}
\label{fig:case20}
\end{figure}

Fig.~\ref{fig:case20} investigates three cases of input, real target and generated target graph by NEC-DGT. The green nodes refer to the uncompromised devices while the red nodes refer to the compromised devices. The width of each edge reflects the distance between two devices. In the first case, both in generated and real target graphs, Devices 4 and 6 are restored to normal, while Device 19 get attacked and is isolated from the other devices. It validates that our NEC-DGT successfully finds the rules of translating nodes and performs like the true confinement process. In the second case, Device 8 propagates the malware to Device 38, which is also modeled by NEC-DGT in generated graphs. In addition, the NEC-DGT not only correctly predicts the nodes attributes, but also discovers the change in edge attributes, e.g. in the third case, most of the connections of compromised Device 10 were cut both in generated and real target graphs.

\subsubsection{Metric-based Evaluation for Molecule Reaction datasets}
In this task, the NEC-DGT is compared to the Weisfeiler-Lehman Difference Network (WLDN)~\cite{jin2017predicting}, which is a graph learning model specially for reaction prediction. Table~\ref{table:mole dataset} shows the performance of our NEC-DGT on the reaction dataset on five metrics, which are the same with the synthetic datasets. The proposed NEC-DGT outperforms both the translation model GT-GAN and the WLDN by 5\% on average.
Though the atoms do not change during reaction, we evaluate the capacity of our NEC-DGT to copy the input node features. As shown in Table ~\ref{table:mole dataset}, The NEC-DGT get the smallest N-MSE and get higher N-R2 than other comparison methods by around 18\%. This shows that our NEC-DGT can deal with a wide range of real-world applications, whether the edges and nodes need change or keep stable.  

\begin{table}[htb]\scriptsize\renewcommand\arraystretch{1.1}
  \centering
  %\captionsetup{font={small}}
   \caption{Evaluation of Generated Target Graphs for Molecule Dataset: N for node attributes, E for edge attributes}\vspace{-0.4cm}
  \begin{tabular}{p{1.1cm}rp{0.6cm}rr|p{1.1cm}r}\\
    \toprule
   Method   & N-MSE   &N-R2    &N-P &N-Sp&Method &E-Acc \\
    \hline
    % &- &- &- &-  \\
    % &- &- &- &-  \\
    IN & 0.0805& 0.46& 0.13&0.12&GT-GAN&0.8687\\
    %DCRNN &	0.0458&	0.08&0.07	&0.04	&WLDN&0.9667\\
    STGCN &	0.0006&	0.98	&\textbf{0.99}&0.97	&WLDN&0.9667\\
    NR-DGT &0.0008	&0.97	&\textbf{0.99}	&\textbf{0.99}&NR-DGT	&0.9918\\
    NEC-DGT &\textbf{0.0004} & \textbf{0.99}&\textbf{0.99} &\textbf{0.99}&NEC-DGT &\textbf{0.9925}\\
    \bottomrule
  \end{tabular}
  \label{table:mole dataset}
\end{table}

\section{Conclusion and Future Work}
This paper focuses on a new problem: multi-attributed graph translation. To achieve this, we propose a novel NEC-DGT consisting of several blocks which translates a multi-attributed input graph to a target graph. To jointly tackle the different types of interactions among nodes and edges, node and edge translation paths are proposed in each block and the graph spectral-based regularization is proposed to preserve the consistent spectral property of graphs. Extensive experiments have been conducted on the synthetic and real-world datasets. Experiment results show that our NEC-DGT can discover the ground-truth translation rules and significantly outperform comparison methods in terms effectiveness. This paper provides a further step of research for graph translation problems in more general scenarios.

\section*{Acknowledgement}
This work was supported by the National Science Foundation grant: \#1755850, \#1841520, \#1907805, Jeffress Trust Award, and NVIDIA GPU Grant.

\bibliographystyle{IEEEtran}
\bibliography{newbib}

\end{document}